\documentclass[10pt,journal,compsoc]{IEEEtran}
\ifCLASSOPTIONcompsoc
  \usepackage[nocompress]{cite}
\else
  \usepackage{cite}
\fi
\ifCLASSINFOpdf
\else
\fi
\begin{document}
\title{Negative Results in Computer Vision:\\ A Perspective}

\author{Ali Borji,~\IEEEmembership{Member,~IEEE}
\IEEEcompsocitemizethanks{\IEEEcompsocthanksitem A. Borji is with the Center for Research in Computer Vision and the Department of Computer Science at the University of Central Florida, Orlando, FL, 32816-2365.\protect\\
E-mail: aborji@crcv.ucf.edu} } % <-this % stops a space


 \markboth{PREPRINT}%
{Shell \MakeLowercase{\textit{et al.}}: Bare Advanced Demo of IEEEtran.cls for IEEE Computer Society Journals}



\IEEEtitleabstractindextext{%
\begin{abstract}
A negative result is when the outcome of an experiment or a model is not what is expected or when a hypothesis does not hold. Despite being often overlooked in the scientific community, negative results are results and they carry value. While this topic has been extensively discussed in other fields such as social sciences and biosciences, less attention has been paid to it in the computer vision community. The unique characteristics of computer vision, particularly its experimental aspect, call for a special treatment of this matter. In this paper, I will address what makes negative results important, how they should be disseminated and incentivized, and what lessons can be learned from cognitive vision research in this regard. Further, I will discuss issues such as computer vision and human vision interaction, experimental design and statistical hypothesis testing, explanatory versus predictive modeling, performance evaluation, model comparison, as well as computer vision research culture. 

\end{abstract}

\begin{IEEEkeywords}
Negative results, All results, Computer vision, Biological vision, Cognitive Sciences, Statistical testing, Statistics.
\end{IEEEkeywords}}

\maketitle

% is not selected <OR> if conference mode is selected - because compsoc
% conference papers position the abstract like regular (non-compsoc)
\IEEEdisplaynontitleabstractindextext

% \fi
\IEEEpeerreviewmaketitle

\ifCLASSOPTIONcompsoc
\IEEEraisesectionheading{\section{Prologue}\label{sec:introduction}}
\else
\section{Introduction}
\label{sec:introduction}
\fi

\IEEEPARstart{N}{egative} findings are results that do not agree with what researchers hypothesize. Such challenging and sometimes inconclusive findings are often discouraged and buried in the drawers and computers. Therefore, the publication record reflects only a tiny slice of the conducted research. In some sense they fabricate the "dark matter" of science. Such findings, however, still hold value. At the very least they can save resources by preventing scientists from repeating the same experiments. Perhaps the main reason for an overwhelmingly high number of negative results not put forward for dissemination is the lack of incentives. Interestingly, some researchers have even argued that most published findings are false~\cite{ioannidis2005most}. Some also claim that hiding negative results is unethical. Nevertheless, negative results have been and continue to be constructive in the advancement of the science (e.g., Michelson-Morley experiment~\cite{shankland1964michelson}).

To answer whether negative results are important in computer vision, should be published, or even if it makes sense to talk about them, first we need to investigate how computer vision research is conducted relative to scientific practices and methodologies conducted in other fields such as social or biological sciences. Computer vision research consists of a mixture of theoretical and experimental research. A small fraction of publications introduce principled theories for vision tasks (e.g., optical flow~\cite{horn1981determining}). A large number of publications report models and algorithms (e.g., for solving the object detection problem) that are more powerful than contending models. Thus, compared to other fields, computer vision is less hypothesis-driven and more practical. Some negative results offer invaluable insights regarding strength and shortcomings of existing models and theories, whereas others provide smart baselines. The emphasis has traditionally been placed on improving existing models in terms of performance over benchmark datasets. While some papers conduct statistical tests, it is not the common practice. As in some other fields, there is a high tendency among computer vision researchers to submit positive results as such results are often considered to be more novel by the reviewers.

Computer vision has its own unique characteristics making it distinct from other fields thereby demanding a specific treatment of negative results. Firstly, vision is an extremely hard problem which has baffled many smart people throughout its history. The complexity of the problem makes it difficult to run controlled experiments and come up with a universal theory of vision. Secondly, often a large number of variables are involved in building vision algorithms and analyzing large scale data. Further, fair comparison of several competing models using multiple scores exacerbates the problem. To address these, it would be very helpful to borrow from other fields (e.g., natural sciences) where experimental design and statistical testing are integral parts of the scientific research. 

The common practice in experimental hypothesis-driven fields (e.g., cognitive science) include carefully formulating a hypothesis, identifying and controlling confounding factors, designing the right stimulus set, collecting high quality data, and performing appropriate statistical tests. These are complicated to perform in computer vision research as often many factors are involved. In particular, statistical analysis becomes very challenging in presence of many parameters and models. This makes it complicated to decide which statistical test is needed or when statistical analysis is critical to conduct. A principled and systematic gauging of the progress (rather than relying on trials and error and luck) helps judge what truly works and what does not and hence steer the research in the right direction. For instance, we might have not given up on neural networks easily if we did more careful rigorous analyses in the past. 

Notice that dealing with negative results is a very controversial topic and still unsettled in many fields. So, do not expect this writing to touch on all of the aspects. Rather, here I try to shed light on some less explored matters and put computer vision in a broader perspective with respect to science in general, and its related fields such as Neuroscience and Cognitive Science, in particular. Indeed, further discussion is needed in the vision community to converge to a consensus regarding treatment of negative results.

In what follows, first I elaborate on science versus engineering and where computer vision fits. I will continue with a comparison of computer and human vision research and how they relate to each other in terms of goals, research methodologies and practices. This is followed by discussions of negative results and statistical analysis in the context of computer vision. Section~\ref{Dissemination} considers the dissemination of negative results. Finally, a wrap up is presented in the epilogue.

\section{Computer Vision: Engineering or Science?}
Let's start with the question of whether computer vision is a scientific or an engineering discipline, or both. Science is concerned with understanding fundamental laws of nature, whereas engineering involves the application of science to create technology, products and services useful for society. Science asks questions about nature, whereas engineers design solutions to problems. 

As a scientific discipline, computer vision is concerned with gaining high-level understanding from digital images, video sequences, views from multiple cameras, or multi-dimensional data. It seeks to automate tasks that the human visual system can do and involves the development of a theoretical and algorithmic basis to achieve automatic visual understanding. Further, it deals with constructing a physical model of the scene (i.e., how the scene is created), how light interacts with the scene, as well as low-, intermediate-, and high-level descriptions of the scene content~\cite{szeliski2010computer}. In other words, the ultimate goal of machine vision is image understanding, the ability not only to recover image structure but also to know what it represents. As a technological and engineering discipline, computer vision seeks to apply its theories and models for the construction of computer vision systems and applications.

Science and engineering are complementary and are beautifully and happily married in computer vision. We have a very solid in-depth scientific understanding of image formation, camera models, depth perception, stereoscopic vision, motion perception, etc. Some engineering applications, among many, include biometrics (robust face and fingerprint recognition), optical character recognition, gesture recognition, motion capture, game playing, structure from motion, image stitching, machine inspection, retail, 3D model building, medical imaging, automotive safety, autonomous cars, assistive systems, and surveillance (in traffic and security). In this respect, computer vision is both theoretical (e.g., optical flow formulation) and experimental (e.g., model replication, parameters tuning, hacks, and tricks).

\section{Computer Vision and Biological Vision}
Vision is a broad interdisciplinary area. Both computer and human vision systems share the same objective which is converting light into useful signals from which accurate models of the physical world can be constructed. This information helps an agent (e.g., be it a robot or a human) live, act, and survive in its environment.

For a long time, human vision research has been concentrated on understanding the principles and mechanisms by which biological visual systems (with higher emphasis on primate vision) operate. This is in essence a reverse engineering (or inverse graphics) task. Likewise, computer vision research seeks a theory and engineering implementation. Despite sharing the same goal, they own unique characteristics. Early human visual sensory mechanisms, including the retina and the Lateral Geniculate Nucleus (LGN), are much more elaborate than current digital cameras (CCD sensors). Neural networks in higher visual areas (e.g., visual ventral stream) accommodate a sophisticated hierarchical processing through cascades of filtering (modeled as convolution), pooling, lateral inhibition, and normalization mechanisms. The result would be a selective and invariant representation of the objects and scenes. This is somewhat akin to what Convolutional Neural Networks (CNNs)~\cite{lecun1998gradient} do. Almost half of the human brain (considered to the the most complex known physical systems and thus a major scientific challenge) is devoted directly or indirectly to vision. The entire brain needs about 20 watts to operate (enough to run a dim light bulb). A processor as smart as the brain requires at least 10 to 20 megawatts of electricity to operate~\cite{watts_ref}. As to processing speed, the brain is still faster than the fastest supercomputers~\cite{speed}. A remarkable capability of human vision is attention (a.k.a active vision) which allows selecting the most relevant and informative part of the massive incoming visual stimulus (at a rate of $10^8$-$10^9$ bits/sec)~\cite{borji2013state,borji2013quantitative}. Both human and computer vision systems have their own biases. Human vision is extremely sensitive to other faces and optical illusions. Similarly, computer vision systems get easily fooled by adversarial examples~\cite{nguyen2015deep}. One thing that we know, almost for sure, is that vision should be solved by a framework that starts from simple feature extractors and builds increasingly more complex features. This is perhaps because the visual world around us is compositional. 
  
There has indeed been a cross-pollination in the two fields (e.g.,~\cite{kruger2013deep,scheirer2014perceptual,pinto2008real,dicarlo2007untangling,medathati2016bio,tan2015benchmarking,fukushima1982neocognitron,borji2014human,vanrullen2017perception,kriegeskorte2015deep,cox2014neural,serre2007robust,ullman2016atoms}). On the one hand, experimental paradigms and psychophysics tools in cognitive vision have been exploited to study the behavior of computer vision algorithms. For example, Parikh and Zitnick~\cite{parikh2011finding} employed the image jumbling paradigm, introduced in~\cite{vogel2006categorization}, to inspect whether some computer vision algorithms capture local or global scene information. Deng et al.~\cite{deng2013fine} used the bubbling paradigm, proposed by Gosselin and Schyns~\cite{gosselin2001bubbles}, to model fine grained object recognition. The rapid (or ultra rapid) serial visual presentation~\cite{potter1975meaning,thorpe1996speed}, has been utilized to investigate the quality of images generated by Generative Adversarial Networks~\cite{denton2015deep}. Vondrick et al.~\cite{vondrick2015learning}, leveraged human recognition biases to improve machine classifiers. On the other hand, computational tools borrowed from computer vision and machine learning have been exploited heavily to formulate hypotheses and designing behavioral and neurophysiological experiments
to understand biological vision. For example, deep convolutional networks have recently been used to study the representational space in the visual ventral stream (e.g.,~\cite{yamins2014performance}). Moreover, a plethora of computer vision, image processing, and machine learning tools have been utilized in biological vision research for the purposes such as stimulus design, studying cues humans might rely on in solving a task, and modeling single neurons and neural networks.

%[[AI computers are no match for human eyes when it comes to recognizing small details, ]]

In terms of performance, while computer vision has made large strides, it is still nowhere near human vision. In general, it seems that computer vision is better than human vision in some restricted tasks where variability is relatively low (e.g., optical character recognition, fingerprint recognition, frontal face recognition, etc). However, it lags far behind in cases where variability is high (e.g., view invariant object recognition~\cite{borji2014defending}). Current state of the art computer vision techniques revolve around deep learning models~\cite{lecun2015deep}, in particular CNNs~\cite{lecun1998gradient}. These models have outperformed traditional techniques on a wide variety of vision problems. It is even believed that CNNs outperform humans on classic hard problems such as scene recognition~\cite{he2015delving}. Nevertheless, while nature has evolved a very efficient robust biological solution to the problem of vision, computer vision is still looking for a general computational theory, let alone efficient physical implementations (e.g., on Silicon).

While both research communities have accumulated a great wealth of knowledge, they are struggling with some common long-standing challenging problems. Perhaps the biggest of all is the invariance problem which is believed to be the "crux of the recognition problem" the "holy grail of vision"~\cite{dicarlo2007untangling}. Humans are remarkably good at recognizing objects under drastic variations (e.g., illumination, rotation, blur, and occlusion) but the mechanisms behind this capacity in biological vision are still unknown. There has been a great deal of research in computer vision to come up with invariant representations. Although the current state of art algorithms (i.e., CNNs) provide partial invariance to some transformations (e.g., translation, rotation, and scale), a principled theory is yet to be developed. This is where cross talks in both fields can be extremely useful. Another challenge, related to the first one, is the role of feedback and top-down modulation in visual processing. The resurgence of deep neural networks has raised the hope that maybe a universal solution to deal with all vision problems is within our reach. This is even more conceivable when we notice that a) CNNs are rooted in the seminal findings of Hubel and Wiesel~\cite{hubel1962receptive}, and b) biological vision systems might be following similar mechanisms to CNNs (e.g.,~\cite{yamins2014performance,serre2007robust,riesenhuber1999hierarchical,anselmi2014representation}).

As for the research practice, different methodologies have been adopted in the two fields, driven by different (and sometimes short term) goals and constraints (e.g., building a technology versus proving a hypothesis). Research in computer vision is traditionally benchmark-driven where algorithms that perform better than others are favored. The emphasis is on improving accuracy. Meanwhile, human vision research is primarily hypothesis-driven. Hypotheses, null and alternative, are carefully formulated, confounding factors are controlled, an appropriate stimulus set is created, behavioral or physiological data is carefully collected, and appropriate statistical tests are conducted. The outcome is valid until proven otherwise. It is the collection of evidences, for or against a hypothesis, that drives the science. Here I would like to bring an example from my own work in eye movement research. A study by Greene and Wolfe~\cite{greene2012reconsidering} reported a failure in predicting an observer's task from his fixations (essentially a negative result) effectively negating the anecdotal finding of Yarbus~\cite{yarbus1967eye}. In a reinvestigation~\cite{borji2014defending}, we repeated the experiment by considering a larger set of parameters over a bigger dataset. To their contrary, we concluded that Yarbus' hypothesis still holds. Several follow-up studies also supported our finding thus reinforcing the original hypothesis. This excerpt from a review that I have received may give you an idea regarding different research methodologies in the two fields: 
"The computer vision literature adopts a sort of \textit{try everything and see what works} approach $\ldots$. But modeling human attention should be more principled than this, and grounded in observed behavioral and neural responses."

There is a discrepancy of opinions on the relationship between computer and biological vision. Some researchers argue that the human visual system provides a most compelling reference model. We can learn much from it as an existence proof and as a great source of inspiration. Therefore, there should be a symbiosis between the two communities as they address the same problem. Some others, to the contrary, argue that since the two systems function under very different constraints, the artificial vision solution does not necessarily need to mimic the biological solution (in reference to the flying). Nonetheless, it seems, to me, that the time is ripe for both communities to learn from each other. A great wealth of biological data (behavioral and neurophysiological) and computational models (e.g., different CNN architectures) are available that should be linked for understanding the visual system and building better models.

\section{Negative Results in Computer Vision}
"Those who cannot remember the past are condemned to repeat it". \hfill   George Santayana

Let me first define some buzzwords before turning the discussion to computer vision. 

Publication bias: coined by Theodore Sterling in 1959~\cite{sterling1959publication}, points to the situation where "publication of research results depends not just on the quality of the research but also on the hypothesis tested, and the significance and direction of effects detected"~\cite{lian2015short}. It is sometimes referred as the "file drawer effect," or "file drawer problem"~\cite{rosenthal1979file}. As the name suggests, results not supporting the original hypotheses (i.e., negative results or Null hypotheses) often end up buried in researchers' file drawers. It is also called the "Positive-results bias" where positive (or successful) results are more likely to be submitted, or accepted than negative or inconclusive results. Therefore, what is published is not the true representative of all results. Perhaps the main reason behind the tendency towards publishing positive results is the intense competition among scientists. The unwritten "publish or perish" rule drives academics to publish interesting high quality papers in large volumes to get more citations and to secure funds to do their research. 

Negative results either go completely unpublished or are somehow turned into positive results through adjustments (e.g., selective reporting, post-hoc reinterpretation, methods alteration, different data analyses, increasing the number of observations). A generic term coined to describe these post-hoc choices is HARKing ("Hypothesizing After the Results are Known"). I will elaborate on this further in the next section.

There are arguments for and against publishing negative results. Let's look at some supporting arguments first. Such results should be part of the scientific record for the sake of completeness. Without them, literature surveys and meta-analyses would be biased. They can save a lot of efforts by preventing researchers to conduct redundant investigations. Further, they motivate critical evaluation, analytical thinking and discussions thus contributing to the intellectual sophistication of the community. Some counter arguments include the followings. Negative results should be less favored due to the scarcity of space. This seems to be no longer an issue in the age of digital publication. Some people argue that negative results can lead to a phenomenon known as the "cluttered office phenomenon" where in an office full of academic papers, it is hard to tell the good ones from the bad ones~\cite{nelson2012let}. This resonates with computer vision research where the field is already replete with an overwhelmingly high volume of publications per year, making separating the wheat from the chaff daunting. 

Notice that negative result is different than no result. No result is a situation where nothing is complete or a work has been done incompletely or incorrectly thus leading to inconclusive or unreliable findings. Examples include a) not having enough participants to do a meaningful analysis, b) not having a control condition for an intervention, c) faulty measurements, or d) inconclusive or wrong statistical analysis. Nevertheless, negative results carry value, although modest. In theory, information from an experiment is not absolutely zero, if done correctly. Even when an experiment gives the exact same result as before (i.e., replication), a higher confidence towards that finding is gained. In some sense, this resembles probability density estimation but over different explanations for a phenomenon. Not every negative result is interesting though. Let me bring an example in the context of computer vision. Consider training a CNN to do a certain task. Often a lot of trickery is involved to properly train a CNN. Now, should one write a paper on the basis that choosing a certain parameter (e.g., training the network for 10 epochs instead of 100) does not lead to a convergence? At the end of the day, what matters is how much a study adds to what is already known, regardless of the sign of the outcome.

Negative results, with a slightly more liberal definition, highlight limitations, failures, or flaws of computer vision models, datasets, or scores. For instance, adversarial images demonstrate situations where CNNs can be easily fooled~\cite{nguyen2015deep}, whereas humans have no trouble recognizing them. In image captioning literature, it has been shown that a nearest neighbor classifier that simply chooses the caption of the most similar image to the test image, outperforms state of the art methods (in 2015~\cite{devlin2015exploring}). In saliency modeling~\cite{borji2013state}, naive baselines such as the average fixation map or a central Gaussian blob outperform several fixation prediction models~\cite{tatler2007central}. In object detection, some works have identified the cases where histogram of oriented gradients~\cite{dalal2005histograms} features fail on certain detection problems~\cite{vondrick2013hoggles}. Smart baselines (e.g., a classifier that picks the label of the most frequent class) define the lower bounds while human performance gives an upper-bound on performance and helps identify the weak links in models. Thus, these are complementary to negative results and help assess the progress and move the field forward. In addition to these, visualization techniques have also proven to be very effective in understanding and evaluating computer vision models (e.g.,~\cite{zeiler2014visualizing,yosinski2015understanding}). 

A related problem here is the issue of replicability. Replicability of findings is believed to be at the heart of empirical sciences~\cite{asendorpf2013recommendations}. As in other fields, computer vision researchers tend not to replicate other people's works for two major reasons. Either it is possible to replicate someone else's work or it is not. In the former case, a reviewer may find your results boring or predictable. In the latter case, you may be accused of not following the right procedure. So, in both case there is a risk factor involved. To mitigate the risks, the community needs to advocate for well-documented solid negative results and educate the reviewers. 

Overall, due to the nature of the computer vision research, in particular from an engineering perspective, the problem of the negative results and replicability in less pressing compared to biological research. Reporting wrong results can be detrimental in natural sciences (e.g., clinical and medical research) because it has important implications (lives are on the line). According to one of my psychologist friends, Elissa Aminoff, "if it (a computer vision algorithm) works, and is working better than everyone else's, it will eventually be made available for everyone to use. If there was a replication issue, it wouldn't get very far. So, I think in computer vision the discussion is more useful for theoretical advancement, rather than for replication issues."

\section{Statistical hypothesis testing}
"What I cannot create, I do not understand". \\ 
\ \ \  \hfill Richard Feynman

How can we tell for sure a result is negative? It makes sense to claim a result negative only when careful rigorous statistical tests have been conducted. In what follows, I highlight some concerns regarding statistical analysis and testing in computer vision.

Scientific research is usually done in two frontiers: experimental and theoretical~\cite{haralick1986computer}. In the theoretical frontier, the pursuit is for a comprehensive theory or principles (often expressed in mathematical forms) to explain visual phenomena (e.g., image formation). In the experimental frontier, researchers utilize two types of approaches: exploratory analysis (pattern mining) and hypothesis testing. In the first approach, they focus on running experiments to gather data in searching for hypotheses. In the second approach, first a well-defined hypothesis is explicitly formulated and then controlled experiments are performed to test its validity. Computer vision researchers use both frontiers although majority of the effort is on building tools and engineering solutions. 

Statistical testing is an integral part of scientific research in various fields (especially experimental ones). Unfortunately, awakening to standard analysis of experimental results has been slow in computer vision and most of analyses are often carried out ad-hoc and heuristically. The input data to computer vision algorithms and estimates produced by them are often noisy. Thus, there is an inherent uncertainty associated with results produced by these algorithms. These uncertainties are expressed in terms of statistical distributions, and distributions' means and covariances. Reporting uncertainty is often neglected in computer vision or is done in a wrong way~\cite{novozin}. Further, many researchers misunderstand confidence intervals and standard error bars~\cite{belia2005researchers,stat-understand}. 

Hypothesis tests are used in determining whether the outcome of a study would lead to a rejection of the null hypothesis for a pre-specified level of significance. The null hypothesis represents what is believed by default, before seeing any evidence. Statistical significance, p-value, is a probability value indicating whether the outcome of an experiment can happen accidentally or not. The smaller the p-value, the larger the significance. If the p-value is less than the required significance level, then we say the null hypothesis is rejected at the given level (e.g., 5\% or 1\%) of significance (i.e., leading to a conclusion). If the p-value is not less than the required significance level, then the test has no result (not conclusive or negative results). In this case the evidence is insufficient to support a conclusion. 

Data dredging, data fishing, data snooping, p-hacking, and HARKing are tricks and ways to tweak data, consciously or unconsciously, such that statistically significant results can be obtained. When talking about this, people often quote Ronald Coase's famous saying "If you torture the data long enough, it will confess". One major flaw is analyzing the data without first devising a specific hypothesis as to the underlying causality. There is a clear distinction between exploratory versus confirmatory analyses. While searching for patterns in data is legitimate, applying a statistical hypothesis tests on the same data is wrong. A simple way to avoid this problem is to form a hypothesis before carrying out significance tests. Notice that the p-value is valid only if you stick to exactly what you had planned to do in advance. Another way is to conduct randomized out-of-sample tests. Here a data set is randomly partitioned into two subsets. One subset is used for formulating a hypothesis and the other is used for testing the hypothesis. Fortunately, this is routinely done in computer vision research. Another flaw in statistical testing is multiple comparisons. If you try large numbers of hypotheses, the chance that one of them may be positive increases. One solution to overcome this is to simply divide the significance criterion ("alpha") by the number of all significance tests conducted during the study. This is known as the Bonferroni correction~\cite{dunnett1955multiple}. Notice that this is a very conservative test. An alpha of 0.05, divided in this way by 100 to account for 100 comparisons, yields a very stringent per-hypothesis alpha of 0.0005. 

One major challenge when designing experiments is dealing with confounding factors (a.k.a confounders or confounding variables). Not controlling the confounding factors can lead to misleading and useless results. Let me clarify this by an example. Assume you aim to investigate the effect of exercise (independent variable) on weight loss (dependent variable). Let's say you collect data from 2n subjects (n male and n female) and conclude that indeed exercise leads to weight loss. Is this a reliable finding? Maybe not, due to several concerns: a) some subjects might have been using drugs so the weight loss could be attributed to that, b) female subjects might have eaten less than male subjects, so gender might be a confounding factor, c) some subjects might have eaten less than others during the course of the experiment, d) some subjects might have eaten immediately after the exercise while other did not, e) some subjects might have been athletes, f) some subjects might have spent less time exercising than others, and so on. In this regard, it is extremely important to understand the difference between correlation and causation. In positive (negative) correlation, dependent and independent variables increase or decrease together (opposite from one another). Causation is when an observed event appears to have caused a second event. In causal dependency, a dependent variable is linked to an experimentally controlled variable. To truly determine causation, the value of a variable should be set independent of other variables (i.e., randomization). While it is often easy to find many correlations in research, identifying causation often requires additional experiments. For example, shoe size correlates with reading level in children but it not the true reason of better reading ability (the true reason might be age or education). Another example is the myth in ancient Germany where people believed that storks deliver babies (See here~\cite{matthews2000storks,storks} for a discussion of this). Notice that these were only few concerns regarding statistical testing. There are, of course, several other factors to be carefully thought about when running experiments and performing statistical testing.

%Let me bring in an example related to computer vision. Let's say you have designed a system that tells whether a scene is captured in China or in the United States. Let's assume you test your model on a dataset that accidentally has people visible in images taken in China while none of the images taken in US contain people. Can we say for sure this model is able to do the task? Not definitively. The reason is that the model might have discovered that the existence of a person determines the location where it was taken. The model may fail when presented with images with no people in them. In this example, randomly sampling the data and scaling up the size of the dataset might mitigate the problem and reduces the bias. 

Let me bring two examples related to computer vision. In the first example, suppose you have designed a system to tell whether a scene is captured in China or in the United States. Further, assume you have trained your model on a dataset that accidentally has people visible in all images taken in China while none of the images taken in US contain people. Is your model able to do the task? Perhaps not! The model might have discovered that the existence of a person in an image determines the location (i.e., an epiphenomenon). Therefore, it may fail when presented with images from China without people in them. In this example, randomly sampling the data and scaling up the size of the dataset can mitigate the problem and may reduces the bias. In the second example, assume you have a model that predicts whether the resolution of an aerial image is low, intermediate or high. Accidentally, your low-, intermediate-, and high resolution train images are taken from areas covered with snow, rock, and vegetation, respectively. It is likely that your model might have learned to classify images from different regions instead of resolution. All in all, since most of the computer vision models are data- driven, it is crucial to understand what aspects of the data they capture.

%All we need for this is to understand correlations. Correlations are not indications of cause and effect. Therefore, predictive models may have no real explanatory power. 

%What we need for this is an understanding of cause and effect. A good explanatory model may have predictive powers.   [[Machine learning tools.]]

%When we think about something we want to know more about, what we want is often a) to explain what happened in the past or is going on right now or b) predict what may happen next. To do this we create models, either as theories or as simulation models. However, the same model can seldom cover both explanation and prediction.

It is becoming increasingly popular to resort to inexpensive crowdsourcing platforms (e.g., Amazon Mechanical Turk~\cite{amt}) to collect annotated data for training supervised data-hungry models~\cite{kovashka2016crowdsourcing}. This has pros and cons. The pros are a) such large scale data provides more statistical power and is rich for statistical hypothesis testing, and b) the learned models have high expressive power (since the stimulus set is a good representative of the real world). There are several cons associated with it as well. Firstly, there is less control over the stimuli. It is often very cumbersome (sometimes impossible) to inspect images to see if they all are appropriate. Secondly, there is less control over the data collection setup. It is hard to tell whether subjects are qualified, they follow the procedure or are actively engaged in the task. Some strategies have been devised to ensure a certain degree of data quality including a) secretly injecting some images for which annotations are already known, b) designing grading tasks (e.g., asking workers to grade each other's work), and c) collecting multiple annotations for every image to reduce noise. Third, due to these uncertainties and noise in the data it sometimes becomes very challenging to conduct meaningful statistical tests. For instance, it is hard to buy the claim that CNNs (e.g., ResNet~\cite{he2015delving}) outperform humans on the ImageNet recognition task. To harness such problems, cognitive vision researchers have traditionally been conducting laboratory experiments where they had (almost) full control over the stimuli and subjects. But that is about to change due to the big data revolution. So, as time goes by, hopefully better ways to deal with these problems will emerge.

It is important to be aware of the difference between explanation and prediction when conducting scientific research. Explanation regards understanding relationships among events and why certain things happen while others do not. Explanatory modeling involves the application of statistical models to test causal explanations and to prove (or disprove) hypotheses. Prediction, on the other hand, regards estimating or forecasting future outcomes. Predictive models may have no real explanatory power and might be merely based on correlations. They often involve applying data mining algorithms to data (i.e., for predicting output value $Y$ from input values $X$). To better understand the difference between the two, consider a coin toss event. While a person without any knowledge of physics can tell for sure  the coin will land on earth, he might not be able to explain the underlying mechanism (i.e., the rule of gravity). According to this demarcation, a majority of computer vision models are predictive models. Both explanatory and predictive modeling can be useful and the choice depends on the purpose and context. The scientific value of predictive modeling, however, has been debated (See~\cite{shmueli2010explain}). Predictive modeling
is often discarded for scientific purposes such as theory building or testing, but has been valued for its applied utility. Please refer to~\cite{shmueli2010explain,yarkoni2016choosing} for an extended discussion on this topic. 

%Although explanatory modeling is commonly used for theory building and testing, predictive modeling is nearly absent in many scientific fields as a tool for developing theory. From conversations with colleagues in various disciplines it appears that 

One last note here is that sometimes it might be somewhat acceptable to tweak parameters, crunch numbers, etcetera in order to build an engineering product, but when it comes to making precise scientific statements, conducting careful statistical tests becomes inevitable.
% acid testing a (model free or model) hypothesis or theory to understand vision from a scientific perspective, then careful statistical testing becomes inevitable (e.g., deep learning models do better than kernel based models for object recognition!).

\section{Dissemination of negative results}
\label{Dissemination}
Dissemination of positive results is often straightforward. How about when results do not support a desired hypothesis or are inconclusive? Here I discuss the issues surrounding communicating negative results. Some journals such as PLOS ONE, Journal of Negative Results in Biomedicine, and Journal of Articles in Support of the Null Hypothesis explicitly welcome negative, null, or inconclusive results. As of now, there is no systematic way of disseminating negative results in computer vision. Not only that, but also commentary papers like this one are often not welcome as journals prefer technical papers.

Computer vision has a unique model of publication. While there are several highly regarded journals (e.g., IEEE PAMI, IJCV, IEEE TIP, CVIU) to publish the results, top-tier conferences are where the real action happens (e.g., CVPR, ICCV, ECCV, NIPS). A large number of papers are submitted to these conferences and get reviewed in a short period of time (around 3 months with the net reviewing period varying from 1 to 2 months). These conferences are very competitive (acceptance rate of around 25\% to 30\%) thus leaving place only for novel, interesting and often positive results. Although, once in a while interesting negative results appear in these conferences, researchers usually do not risk conducting such studies. Some conferences (ICLR and NIPS; publishing some vision papers) have recently adopted an open review system where the communications between the reviewers and the authors are made available to the public. While this does not directly address publishing negative results, it is an effective way of disclosing the hidden chunk of knowledge to the scientific community. Unfortunately, vision conferences have not yet adopted this platform. The reason might be protecting ideas and ongoing efforts.

An important concern in publishing negative results is giving a fair chance to the original authors (especially in cases where published results are questioned) to respond to the counter arguments. Journals seem to be a better venue for such conversations and open debates. Some fields have already devised effective strategies for dealing with this concern. For example, the Journal of Behavioral and Brain Sciences (BBS) invites other scientists to comment on an accepted paper. The paper and the corresponding comments then get published in the same issue of the journal. Journal of Vision and Journal of Vision Research publish commentary and re-analysis papers (sometimes discussing the negative results) as the "letters to the editor". In all of these journals, all material has to go through the peer review process. These practices enrich the scholarly work.

ArXiv and blogs (e.g.,~\cite{tombone}) are two rising venues for publication. Both, however, suffer from a lack of peer review. One advantage of ArXiv is rapid distribution of findings. One drawback is that sometimes papers are early half-baked progress reports often published to claim an idea. Blogs allow personal opinions and discussions in an informal setting (i.e., conversations). Although very interesting, such a venue includes sporadic, noisy thoughts. Nevertheless, occasionally people exploit these venues for communication or settling a matter. For example, I came across an ArXiv paper~\cite{brendel2017comment} debating the results of a previously peer-reviewed published paper~\cite{nayebi2017biologically} which introduced a new biologically-inspired method for mitigating the adversarial perturbations in CNNs. 

One point to stress here is that negative results should go through a thorough review process. Considering the intense race for publication, it only make sense to accept negative results that address problems which are important, challenging, and of high interest to the community. What we certainly do not need is a replete of negative results (or early progress reports) ending up in ArXiv occupying the limited bandwidth of researchers.

One of the most effective habits in computer vision is sharing code and data which has contributed tremendously to the progresses of the field and has been rightfully incentivized by high number of references to such works (similar to benchmark papers). Not only has this habit proven to be extremely useful to deal with replicability issues and speeding up contributions, it also serves as a good model for incentivizing negative results.

\section{Epilogue}

Here I list the main take-away lessons from this work. 

First, negative results that are conclusive, solid, mature, trustworthy, important, interesting, well-documented, and peer-reviewed and come from rigorous investigations should certainly be welcome. Such results can save a lot of efforts by preventing redundant efforts, add to the intellectual richness of the community, promote scholarly culture, and give tremendous insights regarding limits of computer vision models, datasets, and scores. One chief concern here is that hastily-derived and half-baked negative results can be very dangerous and misleading. Notice that it is usually easier to obtain a negative result than a positive result (e.g., making a model work). What is very important, even more than sign, is the validity of the outcome. Overall, negativity towards negative results is counterproductive and such results should be published given that they follow appropriate and sound scientific methodologies.

Second, negative results should be properly and effectively disseminated, incentivized, shared, encouraged and discussed. A culture needs to be built to recognize and embrace such efforts. Emphasizing the statement by Torralba and Efros~\cite{torralba2011unbiased} that "too much value (is given) to winning a particular challenge", negative results as well as smart baselines can be as important as algorithm development or dataset collection and should be given a fair chance to be presented in conferences and journals. To this end, we may need to change the mindset on a larger scale (e.g., funding agencies). Also, negative results should be disseminated in such a way that the original authors can get a chance to respond (in case of replication failure). If possible, it would be more effective to make the comments and discussions available to community. 

Third, statistical testing has been undermined in computer vision and should be taken into account in the future. Several factors need to be carefully taken into account in conducting statistical testing including selection of the appropriate tests, controlling for confounding factors, compensating for multiple comparisons, etc. Statistical testing should be also exploited in model comparisons. This is even more important when models start to saturate on a dataset raising the question of whether a 1\% improvement in accuracy is meaningful. Overall, statistical analysis becomes increasingly more important as computer vision methods hit the market and become prevalent in daily life. Statistical analysis is also critical to support  hypotheses and advance the science of vision. In this regard, students and computer vision researchers should be encouraged to strengthen their knowledge of statistics.

Fourth, multi-disciplinary aspect of computer vision should be emphasized. Negative results and statistical testing is where the field can clearly learn a lot from from other disciplines, biological sciences in particular. To implement this, events that promote such emphasis should be encouraged (e.g., interdisciplinary tutorials and workshops in conferences, publication of interdisciplinary papers, invited talks, etc). In this respect, a synergy between computer and human vision can pay off well in the future. According to Rama Chellappa~\cite{chellappa2012mathematical} "it is counterproductive if one or more groups of researchers to claim that their view of the elephant is the best one. Putting all our effort together, hopefully we can come up with a general solution to the grand problem of vision." This is particularly important in the age of deep learning where biologically-inspired neural networks have demonstrated a great promise.

%The computer vision literature adopts a sort of "try everything and see what works" approach, as evidenced in the PI's review of the relevant modeling work. But modeling human attention should be more principled than this, and grounded in observed behavioral and neural responses.]]

%[[In biology, they often have research method courses; COGS 303 - Research Methods in Cognitive Systems ]]

Fifth, perhaps less related to the main scope of this writing, is a concern on research attitude. Computer vision is enjoying a great phase of expansion and success. It might be even one of the fastest growing areas in computer science, thanks to industry backup. This has brought along a certain culture which is described well by Geman and Geman in "Science in the age of Selfies"~\cite{geman2016opinion}. These authors argue that researchers spend much of their time announcing ideas rather than formulating them. This partly has to be blamed on distractions caused by high information flow, web, social media, etc. Contributing to these, is the fact that researchers are rewarded for publishing more frequently than higher quality which has encouraged researchers to look for "minimum publishable units." This has led to a point that an overwhelmingly high number of publications appear each year making it practically impossible to track the progress even for senior scientists, let alone the newcomers. This concern has been raised by Alan Yuille~\cite{yuille2012computer} "The conference cycle while adding dynamism often leads to a focus on short-term research, an emphasis on 'sound-bytes', and often small progress, improvements in performance on benchmarked datasets -- rather than long-term quality research. This disrupts the balance between short-term research -- picking the low-hanging fruit -- and long-term research which builds the tools to pick the rest. We suggest reestablishing journal publications with rigorous peer review as the 'gold standard' for referencing, for awarding prizes, for faculty appointments, and promotions." We should certainly benefit from relatively less emphasis on quantity of publications and relatively greater emphasis on quality of research. We need to openly and frankly discuss this in the community and find appropriate remedies.

% or


% % use section* for acknowledgment
% \ifCLASSOPTIONcompsoc
%   % The Computer Society usually uses the plural form
%   \section*{Acknowledgments}
% \else
%   % regular IEEE prefers the singular form
% \fi

% I thank two anonymous reviewers, the associate editor,
% and editor for their suggestions and comments
% which improved this manuscript. I express my gratitude
% to many colleagues for invaluable feedback and
% fruitful discussion that have helped me develop the
% explanatory/predictive argument presented in this article.
% I am grateful to Otto Koppius (Erasmus) and
% Ravi Bapna (U Minnesota) for familiarizing me with
% explanatory modeling in Information Systems, for collaboratively
% pursuing prediction in this field, and for
% tireless discussion of this work. I thank Ayala 

\ifCLASSOPTIONcaptionsoff
  \newpage
\fi



%\bibliographystyle{IEEEtran}

%\vspace*{-10pt}
\bibliographystyle{IEEEtran}
	\bibliography{master}

\begin{thebibliography}{10}
\providecommand{\url}[1]{#1}
\csname url@samestyle\endcsname
\providecommand{\newblock}{\relax}
\providecommand{\bibinfo}[2]{#2}
\providecommand{\BIBentrySTDinterwordspacing}{\spaceskip=0pt\relax}
\providecommand{\BIBentryALTinterwordstretchfactor}{4}
\providecommand{\BIBentryALTinterwordspacing}{\spaceskip=\fontdimen2\font plus
\BIBentryALTinterwordstretchfactor\fontdimen3\font minus
  \fontdimen4\font\relax}
\providecommand{\BIBforeignlanguage}[2]{{%
\expandafter\ifx\csname l@#1\endcsname\relax
\typeout{** WARNING: IEEEtran.bst: No hyphenation pattern has been}%
\typeout{** loaded for the language `#1'. Using the pattern for}%
\typeout{** the default language instead.}%
\else
\language=\csname l@#1\endcsname
\fi
#2}}
\providecommand{\BIBdecl}{\relax}
\BIBdecl

\bibitem{ioannidis2005most}
J.~P. Ioannidis, ``Why most published research findings are false,'' \emph{PLos
  med}, vol.~2, no.~8, p. e124, 2005.

\bibitem{shankland1964michelson}
R.~S. Shankland, ``Michelson-morley experiment,'' \emph{American Journal of
  Physics}, vol.~32, no.~1, pp. 16--35, 1964.

\bibitem{horn1981determining}
B.~K. Horn and B.~G. Schunck, ``Determining optical flow,'' \emph{Artificial
  intelligence}, vol.~17, no. 1-3, pp. 185--203, 1981.

\bibitem{szeliski2010computer}
R.~Szeliski, \emph{Computer vision: algorithms and applications}.\hskip 1em
  plus 0.5em minus 0.4em\relax Springer Science \& Business Media, 2010.

\bibitem{lecun1998gradient}
Y.~LeCun, L.~Bottou, Y.~Bengio, and P.~Haffner, ``Gradient-based learning
  applied to document recognition,'' \emph{Proceedings of the IEEE}, vol.~86,
  no.~11, pp. 2278--2324, 1998.

\bibitem{watts_ref}
``http://discovermagazine.com/2009/oct/06-brain-like-chip-may-solve-computers-big-problem-energy/.''

\bibitem{speed}
``https://www.caseyresearch.com/articles/brain-vs-computer.''

\bibitem{borji2013state}
A.~Borji and L.~Itti, ``State-of-the-art in visual attention modeling,''
  \emph{IEEE transactions on pattern analysis and machine intelligence},
  vol.~35, no.~1, pp. 185--207, 2013.

\bibitem{borji2013quantitative}
A.~Borji, D.~N. Sihite, and L.~Itti, ``Quantitative analysis of human-model
  agreement in visual saliency modeling: A comparative study,'' \emph{IEEE
  Transactions on Image Processing}, vol.~22, no.~1, pp. 55--69, 2013.

\bibitem{nguyen2015deep}
A.~Nguyen, J.~Yosinski, and J.~Clune, ``Deep neural networks are easily fooled:
  High confidence predictions for unrecognizable images,'' in \emph{Proceedings
  of the IEEE Conference on Computer Vision and Pattern Recognition}, 2015, pp.
  427--436.

\bibitem{kruger2013deep}
N.~Kruger, P.~Janssen, S.~Kalkan, M.~Lappe, A.~Leonardis, J.~Piater, A.~J.
  Rodriguez-Sanchez, and L.~Wiskott, ``Deep hierarchies in the primate visual
  cortex: What can we learn for computer vision?'' \emph{IEEE transactions on
  pattern analysis and machine intelligence}, vol.~35, no.~8, pp. 1847--1871,
  2013.

\bibitem{scheirer2014perceptual}
W.~J. Scheirer, S.~E. Anthony, K.~Nakayama, and D.~D. Cox, ``Perceptual
  annotation: Measuring human vision to improve computer vision,'' \emph{IEEE
  transactions on pattern analysis and machine intelligence}, vol.~36, no.~8,
  pp. 1679--1686, 2014.

\bibitem{pinto2008real}
N.~Pinto, D.~D. Cox, and J.~J. DiCarlo, ``Why is real-world visual object
  recognition hard?'' \emph{PLoS Comput Biol}, vol.~4, no.~1, p. e27, 2008.

\bibitem{dicarlo2007untangling}
J.~J. DiCarlo and D.~D. Cox, ``Untangling invariant object recognition,''
  \emph{Trends in cognitive sciences}, vol.~11, no.~8, pp. 333--341, 2007.

\bibitem{medathati2016bio}
N.~K. Medathati, H.~Neumann, G.~S. Masson, and P.~Kornprobst, ``Bio-inspired
  computer vision: Towards a synergistic approach of artificial and biological
  vision,'' \emph{Computer Vision and Image Understanding}, vol. 150, pp.
  1--30, 2016.

\bibitem{tan2015benchmarking}
C.~Tan, S.~Lallee, and G.~Orchard, ``Benchmarking neuromorphic vision: lessons
  learnt from computer vision,'' \emph{Frontiers in neuroscience}, vol.~9, p.
  374, 2015.

\bibitem{fukushima1982neocognitron}
K.~Fukushima and S.~Miyake, ``Neocognitron: A self-organizing neural network
  model for a mechanism of visual pattern recognition,'' in \emph{Competition
  and cooperation in neural nets}.\hskip 1em plus 0.5em minus 0.4em\relax
  Springer, 1982, pp. 267--285.

\bibitem{borji2014human}
A.~Borji and L.~Itti, ``Human vs. computer in scene and object recognition,''
  in \emph{Proceedings of the IEEE Conference on Computer Vision and Pattern
  Recognition}, 2014, pp. 113--120.

\bibitem{vanrullen2017perception}
R.~VanRullen, ``Perception science in the age of deep neural networks,''
  \emph{Frontiers in psychology}, vol.~8, 2017.

\bibitem{kriegeskorte2015deep}
N.~Kriegeskorte, ``Deep neural networks: a new framework for modeling
  biological vision and brain information processing,'' \emph{Annual Review of
  Vision Science}, vol.~1, pp. 417--446, 2015.

\bibitem{cox2014neural}
D.~D. Cox and T.~Dean, ``Neural networks and neuroscience-inspired computer
  vision,'' \emph{Current Biology}, vol.~24, no.~18, pp. R921--R929, 2014.

\bibitem{serre2007robust}
T.~Serre, L.~Wolf, S.~Bileschi, M.~Riesenhuber, and T.~Poggio, ``Robust object
  recognition with cortex-like mechanisms,'' \emph{IEEE transactions on pattern
  analysis and machine intelligence}, vol.~29, no.~3, 2007.

\bibitem{ullman2016atoms}
S.~Ullman, L.~Assif, E.~Fetaya, and D.~Harari, ``Atoms of recognition in human
  and computer vision,'' \emph{Proceedings of the National Academy of
  Sciences}, vol. 113, no.~10, pp. 2744--2749, 2016.

\bibitem{parikh2011finding}
D.~Parikh and C.~L. Zitnick, ``Finding the weakest link in person detectors,''
  in \emph{Computer Vision and Pattern Recognition (CVPR), 2011 IEEE Conference
  on}.\hskip 1em plus 0.5em minus 0.4em\relax IEEE, 2011, pp. 1425--1432.

\bibitem{vogel2006categorization}
J.~Vogel, A.~Schwaninger, C.~Wallraven, and H.~H. B{\"u}lthoff,
  ``Categorization of natural scenes: local vs. global information,'' in
  \emph{Proceedings of the 3rd symposium on Applied perception in graphics and
  visualization}.\hskip 1em plus 0.5em minus 0.4em\relax ACM, 2006, pp. 33--40.

\bibitem{deng2013fine}
J.~Deng, J.~Krause, and L.~Fei-Fei, ``Fine-grained crowdsourcing for
  fine-grained recognition,'' in \emph{Proceedings of the IEEE conference on
  computer vision and pattern recognition}, 2013, pp. 580--587.

\bibitem{gosselin2001bubbles}
F.~Gosselin and P.~G. Schyns, ``Bubbles: a technique to reveal the use of
  information in recognition tasks,'' \emph{Vision research}, vol.~41, no.~17,
  pp. 2261--2271, 2001.

\bibitem{potter1975meaning}
M.~C. Potter, ``Meaning in visual search,'' \emph{Science}, vol. 187, no. 4180,
  pp. 965--966, 1975.

\bibitem{thorpe1996speed}
S.~Thorpe, D.~Fize, and C.~Marlot, ``Speed of processing in the human visual
  system,'' \emph{nature}, vol. 381, no. 6582, p. 520, 1996.

\bibitem{denton2015deep}
E.~L. Denton, S.~Chintala, R.~Fergus \emph{et~al.}, ``Deep generative image
  models using a laplacian pyramid of adversarial networks,'' in \emph{Advances
  in neural information processing systems}, 2015, pp. 1486--1494.

\bibitem{vondrick2015learning}
C.~Vondrick, H.~Pirsiavash, A.~Oliva, and A.~Torralba, ``Learning visual biases
  from human imagination,'' in \emph{Advances in neural information processing
  systems}, 2015, pp. 289--297.

\bibitem{yamins2014performance}
D.~L. Yamins, H.~Hong, C.~F. Cadieu, E.~A. Solomon, D.~Seibert, and J.~J.
  DiCarlo, ``Performance-optimized hierarchical models predict neural responses
  in higher visual cortex,'' \emph{Proceedings of the National Academy of
  Sciences}, vol. 111, no.~23, pp. 8619--8624, 2014.

\bibitem{borji2014defending}
A.~Borji and L.~Itti, ``Defending yarbus: Eye movements reveal observers'
  task,'' \emph{Journal of vision}, vol.~14, no.~3, pp. 29--29, 2014.

\bibitem{lecun2015deep}
Y.~LeCun, Y.~Bengio, and G.~Hinton, ``Deep learning,'' \emph{Nature}, vol. 521,
  no. 7553, pp. 436--444, 2015.

\bibitem{he2015delving}
K.~He, X.~Zhang, S.~Ren, and J.~Sun, ``Delving deep into rectifiers: Surpassing
  human-level performance on imagenet classification,'' in \emph{Proceedings of
  the IEEE international conference on computer vision}, 2015, pp. 1026--1034.

\bibitem{hubel1962receptive}
D.~H. Hubel and T.~N. Wiesel, ``Receptive fields, binocular interaction and
  functional architecture in the cat's visual cortex,'' \emph{The Journal of
  physiology}, vol. 160, no.~1, pp. 106--154, 1962.

\bibitem{riesenhuber1999hierarchical}
M.~Riesenhuber and T.~Poggio, ``Hierarchical models of object recognition in
  cortex,'' \emph{Nature neuroscience}, vol.~2, no.~11, pp. 1019--1025, 1999.

\bibitem{anselmi2014representation}
F.~Anselmi and T.~Poggio, ``Representation learning in sensory cortex: a
  theory,'' Center for Brains, Minds and Machines (CBMM), Tech. Rep., 2014.

\bibitem{greene2012reconsidering}
M.~R. Greene, T.~Liu, and J.~M. Wolfe, ``Reconsidering yarbus: A failure to
  predict observers’ task from eye movement patterns,'' \emph{Vision
  research}, vol.~62, pp. 1--8, 2012.

\bibitem{yarbus1967eye}
A.~L. Yarbus, \emph{Eye movements during perception of complex objects}.\hskip
  1em plus 0.5em minus 0.4em\relax Springer, 1967.

\bibitem{sterling1959publication}
T.~D. Sterling, ``Publication decisions and their possible effects on
  inferences drawn from tests of significance—or vice versa,'' \emph{Journal
  of the American statistical association}, vol.~54, no. 285, pp. 30--34, 1959.

\bibitem{lian2015short}
H.~Lian, Y.~Ruan, R.~Liang, X.~Liu, and Z.~Fan, ``Short-term effect of ambient
  temperature and the risk of stroke: a systematic review and meta-analysis,''
  \emph{International journal of environmental research and public health},
  vol.~12, no.~8, pp. 9068--9088, 2015.

\bibitem{rosenthal1979file}
R.~Rosenthal, ``The file drawer problem and tolerance for null results.''
  \emph{Psychological bulletin}, vol.~86, no.~3, p. 638, 1979.

\bibitem{nelson2012let}
L.~D. Nelson, J.~P. Simmons, and U.~Simonsohn, ``Let's publish fewer papers,''
  \emph{Psychological Inquiry}, vol.~23, no.~3, pp. 291--293, 2012.

\bibitem{devlin2015exploring}
J.~Devlin, S.~Gupta, R.~Girshick, M.~Mitchell, and C.~L. Zitnick, ``Exploring
  nearest neighbor approaches for image captioning,'' \emph{arXiv preprint
  arXiv:1505.04467}, 2015.

\bibitem{tatler2007central}
B.~W. Tatler, ``The central fixation bias in scene viewing: Selecting an
  optimal viewing position independently of motor biases and image feature
  distributions,'' \emph{Journal of vision}, vol.~7, no.~14, pp. 4--4, 2007.

\bibitem{dalal2005histograms}
N.~Dalal and B.~Triggs, ``Histograms of oriented gradients for human
  detection,'' in \emph{Computer Vision and Pattern Recognition, 2005. CVPR
  2005. IEEE Computer Society Conference on}, vol.~1.\hskip 1em plus 0.5em
  minus 0.4em\relax IEEE, 2005, pp. 886--893.

\bibitem{vondrick2013hoggles}
C.~Vondrick, A.~Khosla, T.~Malisiewicz, and A.~Torralba, ``Hoggles: Visualizing
  object detection features,'' in \emph{Proceedings of the IEEE International
  Conference on Computer Vision}, 2013, pp. 1--8.

\bibitem{zeiler2014visualizing}
M.~D. Zeiler and R.~Fergus, ``Visualizing and understanding convolutional
  networks,'' in \emph{European conference on computer vision}.\hskip 1em plus
  0.5em minus 0.4em\relax Springer, 2014, pp. 818--833.

\bibitem{yosinski2015understanding}
J.~Yosinski, J.~Clune, A.~Nguyen, T.~Fuchs, and H.~Lipson, ``Understanding
  neural networks through deep visualization,'' \emph{arXiv preprint
  arXiv:1506.06579}, 2015.

\bibitem{asendorpf2013recommendations}
J.~B. Asendorpf, M.~Conner, F.~De~Fruyt, J.~De~Houwer, J.~J. Denissen,
  K.~Fiedler, S.~Fiedler, D.~C. Funder, R.~Kliegl, B.~A. Nosek \emph{et~al.},
  ``Recommendations for increasing replicability in psychology,''
  \emph{European Journal of Personality}, vol.~27, no.~2, pp. 108--119, 2013.

\bibitem{haralick1986computer}
R.~M. Haralick, ``Computer vision theory: The lack thereof,'' \emph{Computer
  Vision, Graphics, and Image Processing}, vol.~36, no. 2-3, pp. 372--386,
  1986.

\bibitem{novozin}
``http://www.nowozin.net/sebastian/blog/how-to-report-uncertainty.html.''

\bibitem{belia2005researchers}
S.~Belia, F.~Fidler, J.~Williams, and G.~Cumming, ``Researchers misunderstand
  confidence intervals and standard error bars.'' \emph{Psychological methods},
  vol.~10, no.~4, p. 389, 2005.

\bibitem{stat-understand}
``http://scienceblogs.com/cognitivedaily/2008/07/31/most-researchers-dont-understa-1/.''

\bibitem{dunnett1955multiple}
C.~W. Dunnett, ``A multiple comparison procedure for comparing several
  treatments with a control,'' \emph{Journal of the American Statistical
  Association}, vol.~50, no. 272, pp. 1096--1121, 1955.

\bibitem{matthews2000storks}
R.~Matthews, ``Storks deliver babies (p= 0.008),'' \emph{Teaching Statistics},
  vol.~22, no.~2, pp. 36--38, 2000.

\bibitem{storks}
``https://priceonomics.com/do-storks-deliver-babies/.''

\bibitem{amt}
``http://www.mturk.com.''

\bibitem{kovashka2016crowdsourcing}
A.~Kovashka, O.~Russakovsky, L.~Fei-Fei, K.~Grauman \emph{et~al.},
  ``Crowdsourcing in computer vision,'' \emph{Foundations and
  Trends{\textregistered} in Computer Graphics and Vision}, vol.~10, no.~3, pp.
  177--243, 2016.

\bibitem{shmueli2010explain}
G.~Shmueli \emph{et~al.}, ``To explain or to predict?'' \emph{Statistical
  science}, vol.~25, no.~3, pp. 289--310, 2010.

\bibitem{yarkoni2016choosing}
T.~Yarkoni and J.~Westfall, ``Choosing prediction over explanation in
  psychology: Lessons from machine learning,'' \emph{FigShare, https://dx. doi.
  org/10.6084/m9. figshare}, vol. 2441878, p.~v1, 2016.

\bibitem{tombone}
``www.computervisionblog.com.''

\bibitem{brendel2017comment}
W.~Brendel and M.~Bethge, ``Comment on" biologically inspired protection of
  deep networks from adversarial attacks",'' \emph{arXiv preprint
  arXiv:1704.01547}, 2017.

\bibitem{nayebi2017biologically}
A.~Nayebi and S.~Ganguli, ``Biologically inspired protection of deep networks
  from adversarial attacks,'' \emph{arXiv preprint arXiv:1703.09202}, 2017.

\bibitem{torralba2011unbiased}
A.~Torralba and A.~A. Efros, ``Unbiased look at dataset bias,'' in
  \emph{Computer Vision and Pattern Recognition (CVPR), 2011 IEEE Conference
  on}.\hskip 1em plus 0.5em minus 0.4em\relax IEEE, 2011, pp. 1521--1528.

\bibitem{chellappa2012mathematical}
R.~Chellappa, ``Mathematical statistics and computer vision,'' \emph{Image and
  Vision Computing}, vol.~30, no.~8, pp. 467--468, 2012.

\bibitem{geman2016opinion}
D.~Geman and S.~Geman, ``Opinion: Science in the age of selfies,''
  \emph{Proceedings of the National Academy of Sciences}, vol. 113, no.~34, pp.
  9384--9387, 2016.

\bibitem{yuille2012computer}
A.~L. Yuille, ``Computer vision needs a core and foundations,'' \emph{Image and
  Vision Computing}, vol.~30, no.~8, pp. 469--471, 2012.

\end{thebibliography}


% \end{IEEEbiography}


% % insert where needed to balance the two columns on the last page with
% %\newpage



%\vfill


% \vspace*{-30pt}

\begin{IEEEbiographynophoto}{Ali Borji}
received the BS and MS degrees in
computer engineering from the Petroleum University
of Technology, Tehran, Iran, 2001 and
Shiraz University, Shiraz, Iran, 2004, respectively.
He received the PhD degree in cognitive
neurosciences from the Institute for Studies in
Fundamental Sciences (IPM) in Tehran, Iran,
2009. He is currently an assistant professor at
Center for Research in Computer Vision, University of Central Florida. His research interests include visual
attention, visual search, machine learning, robotics, neurosciences, and
biologically plausible vision models. He is a member of the IEEE.
\end{IEEEbiographynophoto}

\end{document}